\documentclass[runningheads]{llncs}
\usepackage{graphicx}
\usepackage{adjustbox}
\usepackage{multirow}
\usepackage{color}
\usepackage{xcolor}
\usepackage{booktabs}
\usepackage{amsmath}
\usepackage{amssymb}
\usepackage{subfigure}
\usepackage{multirow}
\usepackage{subfig}
\usepackage{xspace}
\usepackage{booktabs}
\usepackage{makecell}
\usepackage{dblfloatfix}
\usepackage{float}
\usepackage{color}
\usepackage{longtable}
\usepackage{bm}
\usepackage[english]{babel}
\usepackage{ragged2e}
\usepackage{blindtext}
\usepackage{tablefootnote}

\usepackage{makecell}
\usepackage{array}
\usepackage{ tipa } 
\usepackage{amssymb}
\usepackage{tablefootnote} 

\newcommand{\idx}{\ensuremath{\mathtt{Index}}\xspace}
\newcommand{\retriever}{\ensuremath{\mathtt{Retriever}}\xspace}
\newcommand{\qa}{\ensuremath{\mathtt{QAModel}}\xspace}
\newcommand{\task}{\ensuremath{\mathtt{TaskTemp}}\xspace}
\newcommand{\feedback}{\ensuremath{\mathtt{FeedbackTemp}}\xspace}
\newcommand{\concatenation}{\ensuremath{\mathtt{ConcatTemp}}\xspace}
\newcommand{\prompt}{\ensuremath{\mathtt{Prompt}}\xspace}
\newcommand{\llm}{\ensuremath{\mathtt{LLM}}\xspace}

\begin{document}

\title{Enhancing In-Context Learning with Answer Feedback for Multi-Span Question Answering}

\titlerunning{Enhancing In-Context Learning with Answer Feedback for MSQA}
%
\author{Zixian Huang \and Jiaying Zhou \and Gengyang Xiao \and Gong Cheng}
\authorrunning{Z. Huang et al.}
%
\institute{
State Key Laboratory for Novel Software Technology, Nanjing University, China\\
\email{\{zixianhuang,jyzhou,181840264\}@smail.nju.edu.cn, gcheng@nju.edu.cn}
}

\maketitle              
\begin{abstract}
Whereas the recent emergence of large language models (LLMs) like ChatGPT has exhibited impressive general performance, it still has a large gap with 
fully-supervised models on specific tasks such as multi-span question answering. 
Previous researches found that in-context learning is an effective approach to exploiting LLM, by using a few task-related labeled data as demonstration examples to construct a few-shot prompt for answering new questions. 
A popular implementation is to concatenate a few questions and their correct answers through simple templates, informing LLM of the desired output.
In this paper, we propose a novel way of employing labeled data such that it also informs LLM of some undesired output, by extending demonstration examples with feedback about answers predicted by an off-the-shelf model, e.g., correct, incorrect, or incomplete.
Experiments on three multi-span question answering datasets as well as a keyphrase extraction dataset show that our new prompting strategy consistently improves LLM's in-context learning performance.


\end{abstract}
\section{Introduction}
\label{sec:introduction}
Recently, the rise of large language models~(LLMs)~\cite{GPT_3,instruct_GPT,GPT_4} represented by ChatGPT\footnote{https://openai.com/blog/chatgpt} provides a new paradigm for NLP research, which can perform well using only natural language instructions rather than being trained on the target dataset. 
Based on LLMs, many tasks are expected to be more convenient and accessible to users with different needs, including \emph{multi-span question answering}~(MSQA).
MSQA aims to automatically find one-to-many answers at the span level for a given question, which has attracted many in-depth research works~\cite{msqa,tase} based on pre-trained language models~(PLMs), and has broad application scenarios such as medical question answering~\cite{HealthQuestion,CMQA}. 

\begin{figure}
    \centering
    \includegraphics[width=\textwidth]{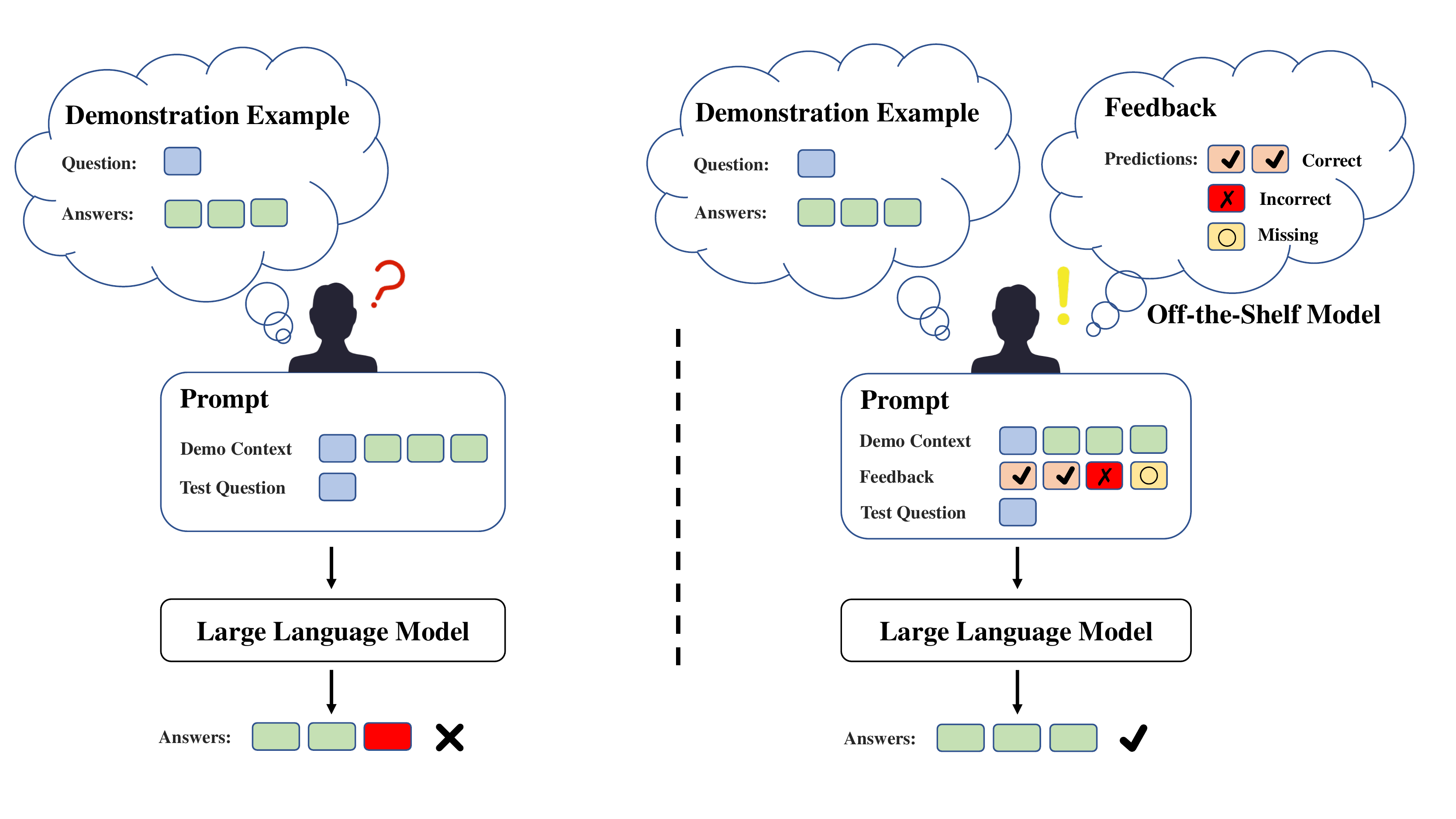}
    \caption{An example of our new prompting strategy (right) compared with the conventional prompting strategy (left). Our strategy first answers the demonstration question using an off-the-shelf model (e.g., based on conventional PLMs) and records the corrected answers as feedback, and then combines demonstration examples with corrected answers to construct a prompt for LLM.}
    \label{fig:idea}
\end{figure}

However, compared with PLMs fine-tuned on the complete training data, LLMs still have a large gap on difficult MSQA datsets~\cite{master_of_none} such as DROP~\cite{DROP,GPT_4}. To address it, \emph{in-context learning}~\cite{in_context_survey} is a promising approach to enhancing the capability of LLMs. The idea of in-context learning is to concatenate the test question with an analogous demonstration context to prompt LLMs to generate answers. As shown in the left half of Figure~\ref{fig:idea}, the demonstration context consists of a few task-related demonstration examples with labeled answers, which can be retrieved from the training set of the target dataset. 

\textbf{Motivation:}
Although existing works have designed a range of approaches for retrieving and exploiting demonstration examples~\cite{KATE,bm25_select,ordering}, the common practice of constructing a demonstration context is still concatenating questions and labeled answers through simple templates. We argue that only showing demonstration questions with correct answers may not guide LLMs to think deeply about demonstration examples, e.g., \emph{lack of reflection on mistakes in problem solving}, which may lead to under-utilization of the labeled answers.



\textbf{Our Work:}
In this paper, we propose to enhance in-context learning with diverse information derived from labeled answers to improve their utilization. Inspired by supervised learning which receives feedback from training loss to update model, we design a novel prompting strategy for LLM to obtain \emph{feedback} information in the form of \emph{corrected answers}.

Specifically, as shown in the right part of Figure~\ref{fig:idea}, this strategy first answers the demonstration question using an off-the-shelf model (e.g., based on conventional PLMs), compares its results with labeled answers, and records the corrected answers as feedback (e.g., correct, incorrect, or missing answers). Then we use both demonstration examples and corrected answers to construct an enhanced prompt for LLM. With this idea, we conducted experiments on three MSQA datasets as well as one keyphrase extraction dataset. The results show that our feedback-based prompting strategy significantly improves the capability of ChatGPT to answer multi-span questions.



\section{Related Work}
\label{sec:rw}

\subsection{Large Language Models}
From GPT-3~\cite{GPT_3} to the latest GPT-4~\cite{GPT_4}, the emergence of powerful LLMs in recent years has triggered new thinkings and paradigms in NLP research. LLMs perform various downstream tasks using only text instructions, have matched state-of-the-art results in many tasks including machine translation~\cite{good_translator} and relation extraction~\cite{GPT_RE}, and have influenced a range of domain applications such as education~\cite{ChatGPT_Exam} and medical writing~\cite{ChatGPT_Medical}. Despite the great success of LLMs, studies have also reported that it still has shortcomings in specific tasks~\cite{ChatGPT_General_Purpose_eval,Multitask_Evaluation_Reasoning} and has a large gap in handling difficult tasks compared with PLM-based methods~\cite{master_of_none}. 

In particular, question answering~(QA) is a task with long-term research and is faced with various challenges. The performance of LLMs on QA has received extensive attention. 
Some analytical works reported that LLMs have many limitations in QA tasks, including insufficient stability~\cite{complex_questions}, poor performance on newly released datasets~\cite{eval_reasoning}, and suffering from hallucinations~\cite{Multitask_Evaluation_Reasoning}.
Based on empirical observations, some works designed methods to improve the performance of LLMs on specific QA tasks such as commonsense QA~\cite{eval_commonsense}, open-domain QA~\cite{Self-Prompting}, and multi-document QA~\cite{multi-doc}. 

However, as an important and realistic QA task, \emph{Multi-Span QA~(MSQA) currently lacks dedicated research based on LLMs, whose performance on this task remains unclear}. In this paper, we propose and evaluate a novel strategy for effectively adapting LLMs to the MSQA task.


\subsection{In-Context Learning}
With the development of LLMs, in-context learning~\cite{in_context_survey} has also received extensive attention in recent years. Some research works studied it from the perspective of demonstration formatting, proposing template engineering to construct better human-written or automatically generated prompts~\cite{prompt_engine,auto_prompt}. Some other methods enhanced in-context learning by selecting better demonstration examples, searching for the best ordering of demonstration examples~\cite{ordering}, or using the KNN algorithm with lexical~\cite{bm25_select} or semantic~\cite{KATE} features to dynamically retrieve demonstration examples for each question. 

The usage of labeled answers in the above methods is to append them to the question using some simple templates, which leads to potential under-utilization of labeled answers. The work most similar to ours is~\cite{GPT_RE}, which feeds demonstration examples to LLM to obtain a clue about the gold labels in a given document in a relation extraction task. However, the clue generated by LLM often contains mistakes, which also causes some loss of label information, and it is very expensive to interact every demonstration example with LLM. By contrast, in this paper, \emph{we obtain answer feedback by comparing the prediction results on the demonstration example with the labeled answers, and use it to enrich in-context learning with more insightful information obtained from the corrected answers}.

\section{Approach}
\label{sec:approach}
Given a question $Q$ and a reference document $D$, the goal of MSQA is to generate a set of $n$~answers $\mathcal{A}=\{A_1, \ldots, A_n\}$, where $A_i$ is a span-level text that may be either present in $D$ or absent in $D$. Let $\mathcal{T}=\{[D^T_1, Q^T_1, \mathcal{A}^T_1], \ldots \}$  be a set of labeled examples, i.e., the set of all the available question-document-answers triples from which demonstration examples can be selected for in-context learning, e.g., the training set of a MSQA dataset.

\begin{figure}
    \includegraphics[width=1.0\textwidth]{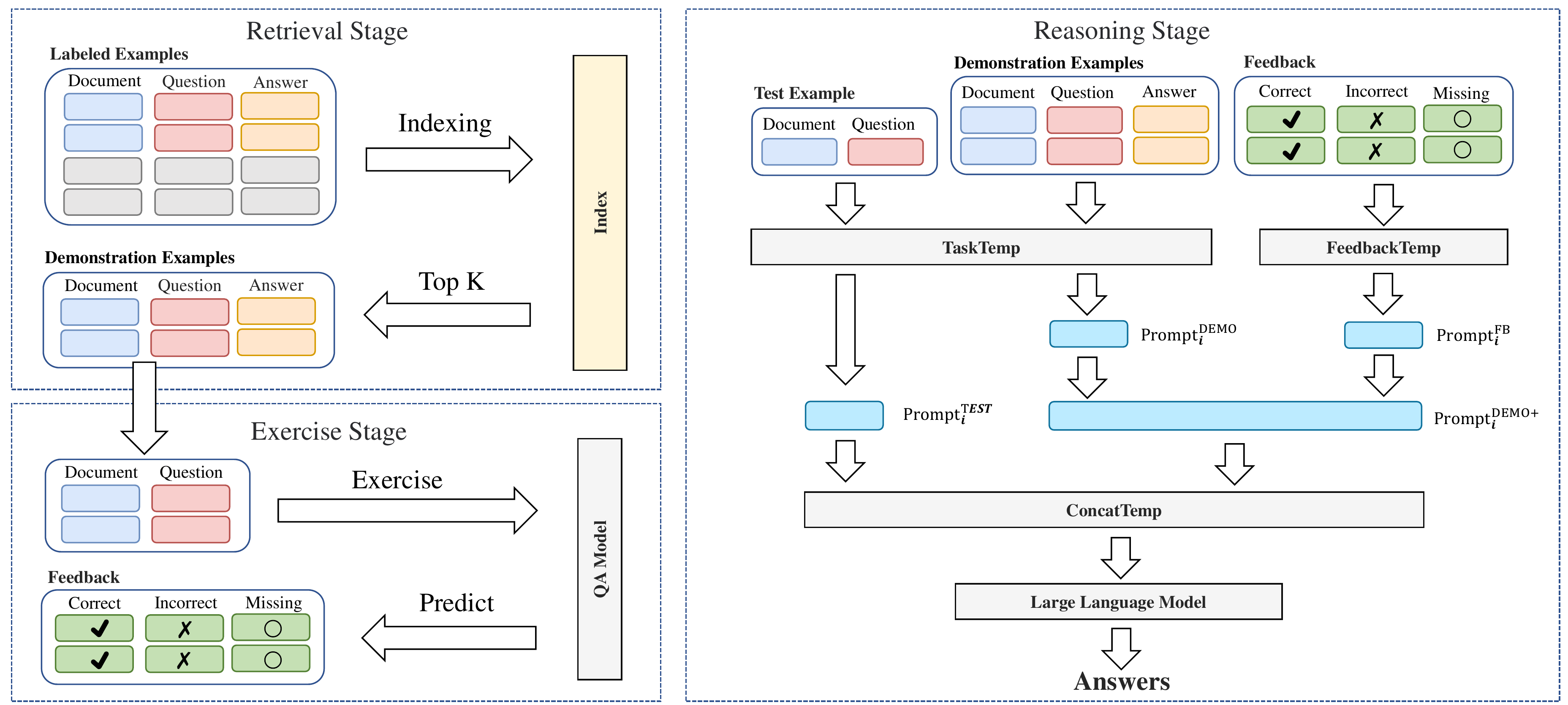}
    \caption{An overview of our prompting strategy, which includes a retrieval stage
searching for relevant demonstration examples, an exercise stage for producing feedback, and a reasoning stage for in-context learning with feedback.}
    \label{fig:framework}
\end{figure}

Figure~\ref{fig:framework} gives an overview of our strategy, which includes a retrieval stage searching for relevant demonstration examples, an exercise stage for producing feedback, and a reasoning stage for in-context learning with feedback.

\subsection{Retrieval Stage}
We first search for a few relevant demonstration examples for test question $Q$ from the labeled examples set $\mathcal{T}$. To this end, a question index $ \mathcal{I}$ is built for each question $Q^T_i$ in $\mathcal{T}$, and a retrieval module is executed to obtain the set $\mathcal{E}$ of top-$k$ relevant labeled examples:
\begin{align}
\label{eq:retrieval}
\small
\begin{split}
    \mathcal{I} &= \idx(\mathcal{T}) \\
    \mathcal{E} &= \retriever(Q, \mathcal{I}),
    \text{where} \quad \mathcal{E} \subset \mathcal{T} \,,
\end{split}
\end{align}

\noindent where $\idx(\cdot)$ and $\retriever(\cdot, \cdot)$ are indexing and retrieval functions, respectively, and we realize them using an inverted index and BM25 in our experiments. $\mathcal{E}=\{[D^E_1, Q^E_1, \mathcal{A}^E_1], \ldots, [D^E_k, Q^E_k, \mathcal{A}^E_k]\}$ is the selected demonstration examples set with size $k$.

\subsection{Exercise Stage}
Then we regard the selected demonstration examples $\mathcal{E}$ as exercises to predict their answers and extend them with corrected answers as feedback. The set of predicted answers $\mathcal{A}^P_i$ for each demonstration question $Q^E_i$ is obtained as follows:
\begin{align}
\label{eq:exercise}
\small
\begin{split}
    \mathcal{A}^P_i &= \qa(D^E_i, Q^E_i) \,,
\end{split}
\end{align}
where $\qa(\cdot, \cdot)$ is an off-the-shelf MSQA model (e.g., a conventional MSQA method based on PLMs), and $\mathcal{A}^P_i=\{ A^P_1,\ldots,A^P_m\}$ is the predicted answers set with size $m$. 

Next, the predicted answers set $\mathcal{A}^P_i$ is compared with the labeled answers set $\mathcal{A}^E_i$ to obtain feedback about the predicted answers. The feedback consists of three parts: the correctly predicted set $\mathcal{A}^C_i$, the incorrectly predicted set $\mathcal{A}^I_i$, and the unpredicted (i.e., missing) set $\mathcal{A}^M_i$, satisfying that $|\mathcal{A}^C_i| + |\mathcal{A}^I_i| = m$ and $|\mathcal{A}^C_i| + |\mathcal{A}^M_i| = n$.

\subsection{Reasoning Stage}

\begin{table}[h]
\scriptsize
\caption{Prompting templates used for MSQA and Keyphrase Extraction (KE)}\label{tab:prompt}
\renewcommand\arraystretch{1.1}
\begin{adjustbox}{center}
\begin{tabular}{p{1.1cm}|p{1.9cm}|p{9.0cm}}

\Xhline{3\arrayrulewidth}
\makecell[c]{Task} & \makecell[c]{Function} & \makecell[c]{Templates}\\

\hline
\multirow{9}{*}{\centering \shortstack{\hspace{0.4em}MSQA \\ \hspace{0.4em}\& KE}} & \multirow{9}{*}{\shortstack{\hspace{0.1em}FeedbackTemp\\ \hspace{0.1em}($\cdot,\cdot,\cdot$)}} & Here are some \textbf{correct} answers (or present/absent keyphrases) responded by other AI model:\\
& & 1. \textbf{[\textit{CORRECT1}]}; 2. \textbf{[\textit{CORRECT2}]}; ...\\
& & Here are some \textbf{incorrect} answers (or present/absent keyphrases) responded by other AI model:\\
& & 1. \textbf{[\textit{INCORRECT1}]}; 2. \textbf{[\textit{INCORRECT2}]}; ...\\
& & Here are some answers (or present/absent keyphrases) \textbf{missed} by other AI model:\\
& & 1. \textbf{[\textit{MISS1}]}; 2. \textbf{[\textit{MISS2}]}; ...\\

\Xhline{3\arrayrulewidth}
\multirow{10}{*}{\centering \shortstack{\hspace{0.3em}MSQA}} & \multirow{5}{*}{\centering \shortstack{\hspace{1.05em}TaskTemp\\ \hspace{1.05em}($\cdot,\cdot,\cdot$)}} & Reading the passage: \textbf{[\textit{DOCUMENT}]} \\
& & Extract spans from the above passage to answer the question: \textbf{[\textit{QUESTION}]} \\
& & Answer as a list e.g. 1. answer1; 2. answer2 \\
& & Answer: 1. \textbf{[\textit{ANS1}]}; 2. \textbf{[\textit{ANS2}]}; ...\\
\cline{2-3}
& \multirow{5}{*}{\shortstack{\hspace{0.6em}ConcatTemp\\ \hspace{0.6em}($\cdot,\cdot$)}} & Example1: \textbf{[\textit{DEMO CONTEXT1}]}\\
& & Example2: \textbf{[\textit{DEMO CONTEXT2}]} \\
& & ... \\
& & Then, answer me a question like the above examples: \\
& & \textbf{[\textit{TEST QUESTION}]} \\
\Xhline{3\arrayrulewidth}

\multirow{9}{*}{\shortstack{\hspace{1.0em}KE}} & \multirow{4}{*}{\centering \shortstack{\hspace{1.05em}TaskTemp\\ \hspace{1.05em}($\cdot,\cdot,\cdot$)}} & Reading the passage: \textbf{[\textit{DOCUMENT}]} \\
& & Extract present (or Generate absent) keyphrases from the above passage: \\
& & Response as a list e.g. 1. keyphrase1; 2. keyphrase2 \\
& & Keyphrases: 1. \textbf{[\textit{KEYPHRASE1}]}; 2. \textbf{[\textit{KEYPHRASE2}]}; ... \\
\cline{2-3}
& \multirow{5}{*}{\shortstack{\hspace{0.6em}ConcatTemp\\ \hspace{0.6em}($\cdot,\cdot$)}} & Example1: \textbf{[\textit{DEMO CONTEXT1}]}\\
& & Example2: \textbf{[\textit{DEMO CONTEXT2}]} \\
& & ... \\
& & Then, extract present (or generate absent) keyphrases like the above cases:\\
& & \textbf{[\textit{TEST QUESTION}]} \\



\Xhline{3\arrayrulewidth}
\end{tabular}
\end{adjustbox}
\end{table}

After obtaining the answer feedback, an extended demonstration context is constructed from $\mathcal{E}$ and the feedback. For each demonstration example, we use a task description template to construct demonstration context $\prompt_i^\text{DEMO}$, use a feedback template to construct feedback context $\prompt_i^\text{FB}$, and the expended demonstration context $\prompt_i^\text{DEMO+}$ is constructed by concatenating $\prompt_i^\text{DEMO}$ and $\prompt_i^\text{FB}$:
\begin{align}
\small
\begin{split}
    \prompt_i^\text{DEMO} &= \task(D^T_i, Q^T_i, \mathcal{A}^T_i)\\
    \prompt_i^\text{FB} &= \feedback(\mathcal{A}^C_i, \mathcal{A}^I_i, \mathcal{A}^M_i)\\
    \prompt_i^\text{DEMO+} &= [\prompt_i^\text{DEMO};\prompt_i^\text{FB}] \,,
\end{split}
\end{align}
where $\task(\cdot, \cdot, \cdot)$ and $\feedback(\cdot, \cdot, \cdot)$ are two template filling functions. The details of the templates can be found in Table~\ref{tab:prompt}.

For the test question $Q$, we construct test context using the same task description template but set the answers to an empty set:
\begin{align}
\small
\begin{split}
    \prompt_i^\text{TEST} &= \task(D, Q, \varnothing) \,.
\end{split}
\end{align}

Finally, we use a concatenation template to construct the complete prompt and feed it into LLM: 
\begin{align}
\small
\begin{split}
    \prompt &= \concatenation(\{\prompt_i^\text{DEMO+}, \dots\}, \prompt_i^\text{TEST}) \\
    A^\text{LLM}&=\llm(\prompt)\,,
\end{split}
\end{align}
where $\concatenation(\cdot, \cdot)$ is a template filling function detailed in Table~\ref{tab:prompt}, and $A^\text{LLM}$ is a text answer returned by LLM. Since the instruction in the prompt requires LLM to answer in the form of a list, we can easily parse the text into multiple span-level answers to the test question.



\section{Experimental Setup}

We refer to our approach as \textbf{FBPrompt}.

\subsection{Datasets}

We compared FBPrompt with baselines on three MSQA datasets: \textbf{MultispanQA}~\cite{DROP}, \textbf{QUOREF}~\cite{quoref}, and \textbf{DROP}~\cite{msqa}. 
Since the test set of them is hidden, we used the official development set as our test set. In addition, we used a keyphase extraction dataset \textbf{Inspec}~\cite{inspec}, which has a similar format to MSQA, with one document input and multiple span-level outputs,  but without question. Considering the experimental cost, we only randomly sampled 500 samples for evaluation from QUOREF and DROP. Table~\ref{table:datasetstatistics} shows some statistics about these datasets.

\begin{table}
\caption{Dataset statistics. Present Labels (\%) indicates the percentage of answers in MSQA datasets or percentage of keyphrases in keyphrase extraction datasets that explicitly appear in the document.}
\resizebox{\textwidth}{!}{
\begin{tabular} {l|>{\hspace{0.2cm}}c<{\hspace{0.2cm}}|>{\hspace{0.2cm}}c<{\hspace{0.2cm}}|>{\hspace{0.2cm}}c<{\hspace{0.2cm}}|>{\hspace{0.2cm}}c<{\hspace{0.2cm}}|>{\hspace{0.2cm}}c<{\hspace{0.2cm}}}
\Xhline{2\arrayrulewidth}
    Dataset & Type &\#~Test & \#~Used & Present Labels~(\%) & Avg.~\# Answers \\
    \hline
    MultiSpanQA\cite{msqa} & MSQA & 653 & 653 & 100 & 2.89\\
    QUOREF\cite{quoref} & MSQA & 2537 & 500 & 100 & 1.14\\
    DROP\cite{DROP} & MSQA & 9,622 & 500 & 73.03 & 1.09\\
    INSPEC\cite{inspec} & KP & 500 & 500 & 26.42 & 2.48\\
    
\Xhline{2\arrayrulewidth}
\end{tabular}
}

\label{table:datasetstatistics}
\end{table}

\subsection{Baselines}
We compared FBPrompt with  five popular usages of LLM as follows:

\textbf{Zero-shot} prompts LLM only using handle-written instructions without demonstration examples.

\textbf{Random} Sampling randomly selects $k$ demonstration examples from the training set for each test question to construct prompt as done in~\cite{KATE}.

\textbf{BM25} calculates lexical similarity between questions to obtain top-$k$ relevant demonstration examples for each test question. It can be viewed as a simplified version of our FBPrompt---without using answer feedback.

\textbf{KATE}~\cite{KATE} uses KNN algorithm selecting $k$ demonstration examples with highest semantic similarity score for each test question. We implemented it based on dense passage retrieval~\cite{dpr}.

\textbf{Label-induced} Reasoning~\cite{GPT_RE} feeds labeled answers, the question, and the document to LLM to obtain a clue about the relation between question and answers. We implemented it using the same BM25 results as our FBPrompt.



\subsection{Evaluation Metrics}
We evaluated on each dataset using their official metrics~\cite{msqa,DROP,quoref,kpeval}. For MultiSpanQA, we used Exact Match F1~(\textbf{EM}) 
and Partial Match F1~(\textbf{PM}).
For QUOREF and DROP, 
we used Exact Match Global~(\textbf{$\text{EM}_\text{G}$}) 
and F1 score~(\textbf{F1}). 
For INSPEC, we used macro-averaged \textbf{F1@5} and \textbf{F1@M}.



\subsection{Implementation Details}
We used OpenAI official API\footnote{https://platform.openai.com/} with the model gpt-3.5-turbo-0301 for all our experiments. We used the T5-base~\cite{T5} model as our off-the-shelf model in FBPrompt. For the keyphrase extraction task, we performed extraction of present keyphrases and generation of absent keyphrases in two independent steps with two slightly different instructions as show in Table~\ref{tab:prompt}. Unless otherwise specified, we set $k=3$, i.e., FBPrompt and all the few-shot baselines used three demonstration examples.
\section{Experimental Results}

\subsection{Comparison with Baselines}
\begin{table}
\centering
\caption{Main results on MSQA. The best results are in bold. \textsuperscript{‡} indicates the results reported in \cite{GPT_Keyphrase}. \textsuperscript{$\ast$} indicates that the results are not completely comparable due to the difference in test data.}
\label{table:main_results}

\begin{tabular}{lcc|cc|cc|cc|cc}

\Xhline{3\arrayrulewidth}
\multirow{3}{*}{} & \multicolumn{2}{c|}{MultiSpanQA} & \multicolumn{2}{c|}{QUOREF} & \multicolumn{2}{c|}{DROP} & \multicolumn{4}{c}{INSPEC} \\
\cline{2-11}
& \multirow{2}{*}{EM} & \multirow{2}{*}{PM} & \multirow{2}{*}{EM$_\text{G}$} & \multirow{2}{*}{F1} & \multirow{2}{*}{EM$_\text{G}$} & \multirow{2}{*}{F1} & \multicolumn{2}{c|}{Present} & \multicolumn{2}{c}{Absent}\\
\cline{8-11}
& & & & & & & F1@5 & F1@M & F1@5 & F1@M \\
\Xhline{2\arrayrulewidth}

SOTA & \textbf{73.13}\textsuperscript{$\ast$} & \textbf{83.36}\textsuperscript{$\ast$} & \textbf{80.61}\textsuperscript{$\ast$}
 & \textbf{86.70}\textsuperscript{$\ast$} & \textbf{84.86}\textsuperscript{$\ast$} & \textbf{87.54}\textsuperscript{$\ast$} & 0.401\textsuperscript{‡} & 0.476\textsuperscript{‡} & 0.030\textsuperscript{‡} & 0.041\textsuperscript{‡}
\\
\Xhline{2\arrayrulewidth}
Zero-shot &39.47&68.14&33.60&51.07& 5.81 & 17.25 & 0.298\textsuperscript{‡} & 0.417\textsuperscript{‡} & 0.016\textsuperscript{‡} & 0.030\textsuperscript{‡}
\\
Random & 58.62 & 80.62 & 71.40 & 80.25 & 47.70 & 60.53 & 0.401 & 0.472 & 0.033 & 0.051
\\
Label-induced & 54.56 & 76.99 & 64.40 & 71.96 & 12.63 & 16.47 & 0.115 & 0.135 & 0.009 & 0.013 \\

KATE & 60.78 & 81.51 & 73.00 & 79.76 & 50.90 & 60.69 & 0.399 & 0.468 & 0.026 & 0.038
\\
BM25 & 61.33 & 81.63 & 70.80 & 79.00 & 58.40 & 65.93 & 0.405 & 0.470 & 0.029 & 0.051
\\
\Xhline{2\arrayrulewidth}
FBPrompt & 64.60 & 83.11 & 73.60 & 80.55 & 62.00 & 69.11 & \textbf{0.425} & \textbf{0.499} & \textbf{0.034} & \textbf{0.055}
\\

\Xhline{3\arrayrulewidth}
\end{tabular}


\end{table}

In Table~\ref{table:main_results},  FBPrompt significantly outperforms previous LLM-based methods on all metrics in the four datasets. In particular, compared with BM25 which uses the same demonstration examples as ours,  FBPrompt exceeds it by a large lead, thus exhibiting the performance brought by our proposed answer feedback.  

We also show the state-of-the-art~(SOTA) results reported by other papers using fully-supervised fine-tuned models: they are \cite{liquid} for MultiSpanQA, \cite{tase} for QUOREF, \cite{opera} for DROP, and \cite{GPT_Keyphrase} for INSPEC. Although the experimental results on the three MSQA datasets are not directly comparable due to inconsistent test data, it can be found that LLM-based models are still weaker than the fully-supervised models, but performs relatively well on the keyphrase extraction dataset INSPEC. FBPrompt closes the gap to SOTA on MSQA and achieves new SOTA results on the INSPEC dataset.


\subsection{The Effectiveness of Different Feedback}
 We compare FBPrompt with the method using only one type of feedback to analyze whether all the three types of feedback bring benefits. The results reported in Table~\ref{table:ablation_study1} reveal that each part of the feedback has a effect in improving the performance of LLM. In particular, using only correct answers leads to the largest loss compared with using only incorrect or missing answers, which shows that negative feedback has the largest benefit to LLM.

\begin{table}
\centering
\caption{Effectiveness of different feedback. The best results are in bold.}
\label{table:ablation_study1}
\begin{tabular}{lcc|cc|cc|cc|cc}

\Xhline{3\arrayrulewidth}
\multirow{3}{*}{} & \multicolumn{2}{c|}{MultiSpanQA} & \multicolumn{2}{c|}{QUOREF} & \multicolumn{2}{c|}{DROP} & \multicolumn{4}{c}{INSPEC} \\
\cline{2-11}
& \multirow{2}{*}{EM} & \multirow{2}{*}{PM} & \multirow{2}{*}{EM$_\text{G}$} & \multirow{2}{*}{F1} & \multirow{2}{*}{EM$_\text{G}$} & \multirow{2}{*}{F1} & \multicolumn{2}{c|}{Present} & \multicolumn{2}{c}{Absent}\\
\cline{8-11}
& &&&&&& F1@5 & F1@M & F1@5 & F1@M \\
\Xhline{2\arrayrulewidth}

FBPrompt & \textbf{64.60} & \textbf{83.11} & \textbf{73.60} & \textbf{80.55} & \textbf{62.00} & \textbf{69.11} & \textbf{0.425} & \textbf{0.499} & \textbf{0.034} & \textbf{0.055}
\\
~- only correct & 62.70 & 82.75 & 71.40 & 79.69 & 58.40 & 65.60 & 0.401 & 0.463 & 0.027 & 0.046
\\
~- only incorrect & 62.93 & 82.97 & 72.40 & 80.23 & 60.20 & 67.92 & 0.417 & 0.490 & 0.030 & 0.048
\\
~- only missing & 63.48 & 82.90 & 72.80 & 79.75 & 61.20 & 68.80 & 0.416 & 0.480 & 0.027 & 0.046
\\

\Xhline{3\arrayrulewidth}

\end{tabular}
\end{table}

\begin{figure}[b]

  \centering
  \subfigure[MultiSpanQA]{\includegraphics[width=0.23\textwidth]{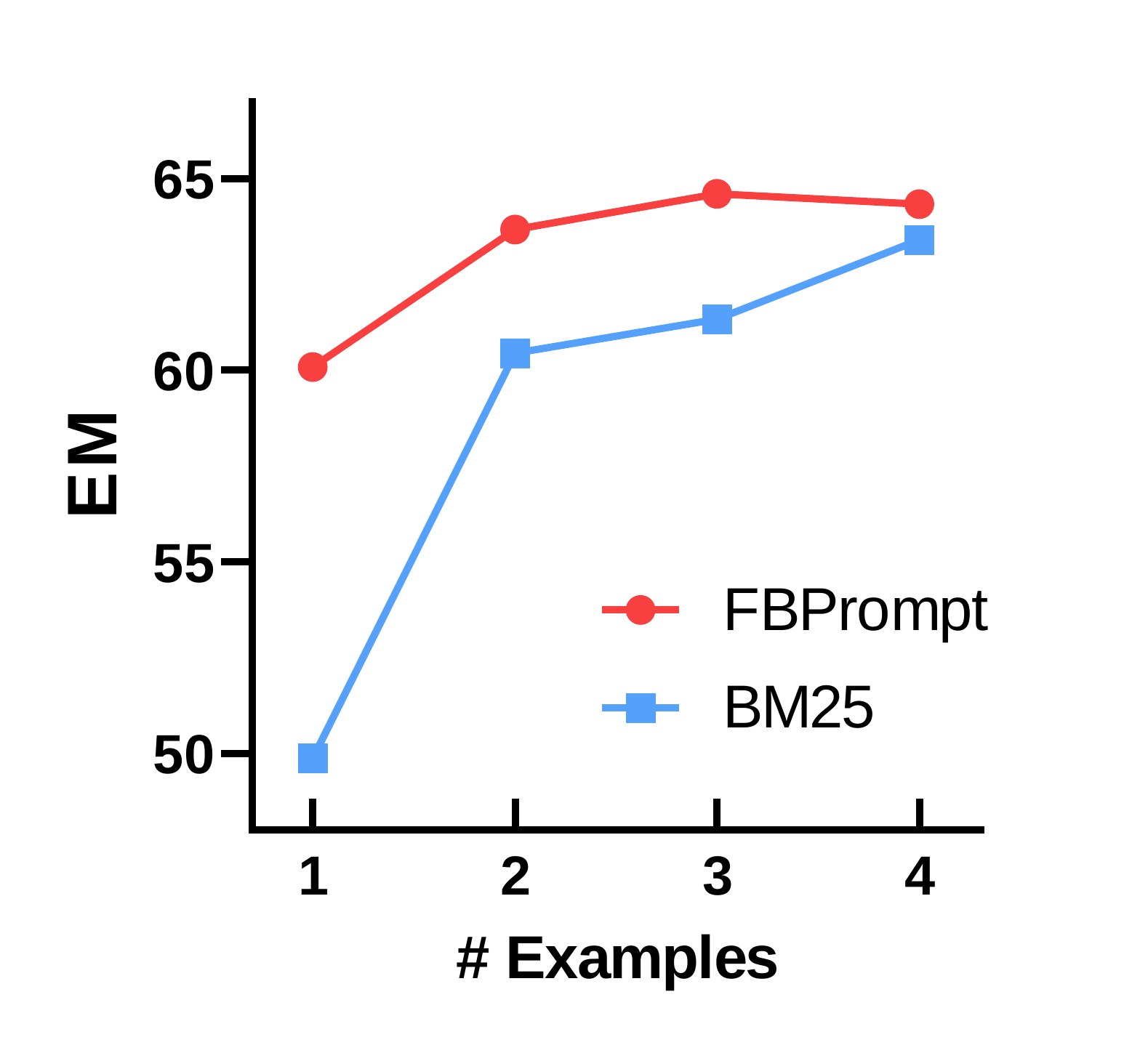}} 
  \subfigure[QUOREF]{\includegraphics[width=0.23\textwidth]{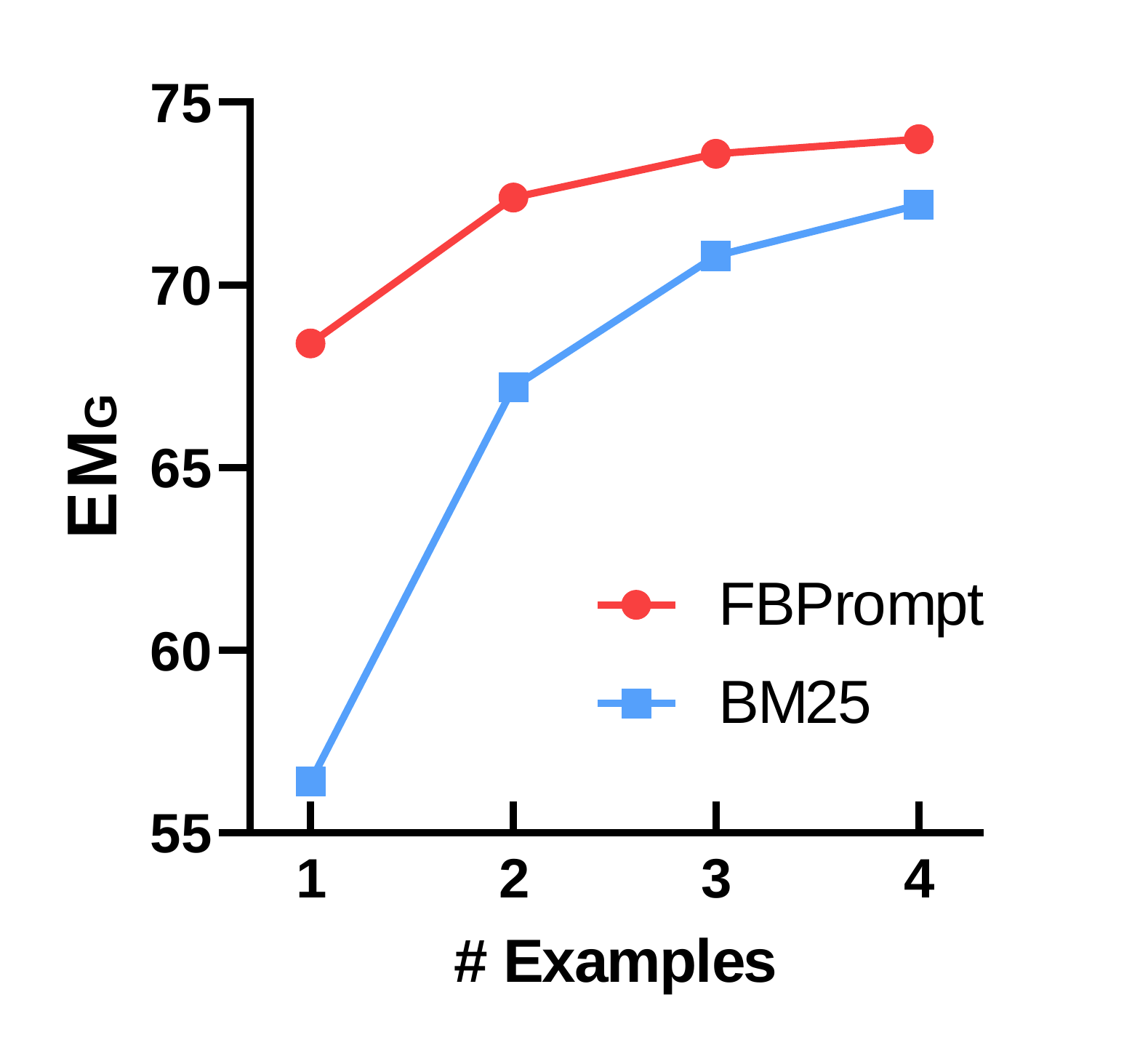}}
  \subfigure[DROP]{\includegraphics[width=0.23\textwidth]{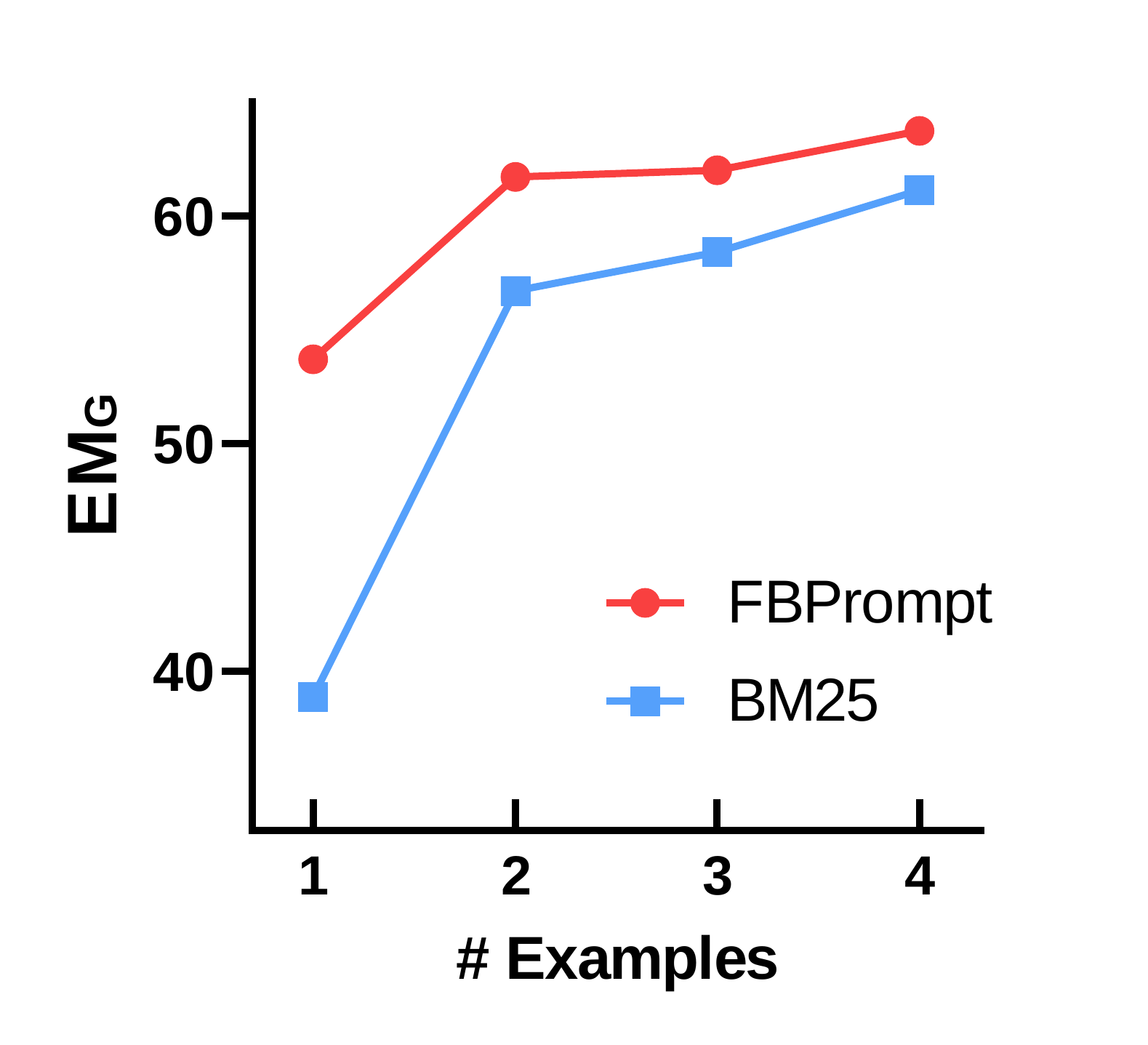}}
  \subfigure[INSPEC]{\includegraphics[width=0.23\textwidth]{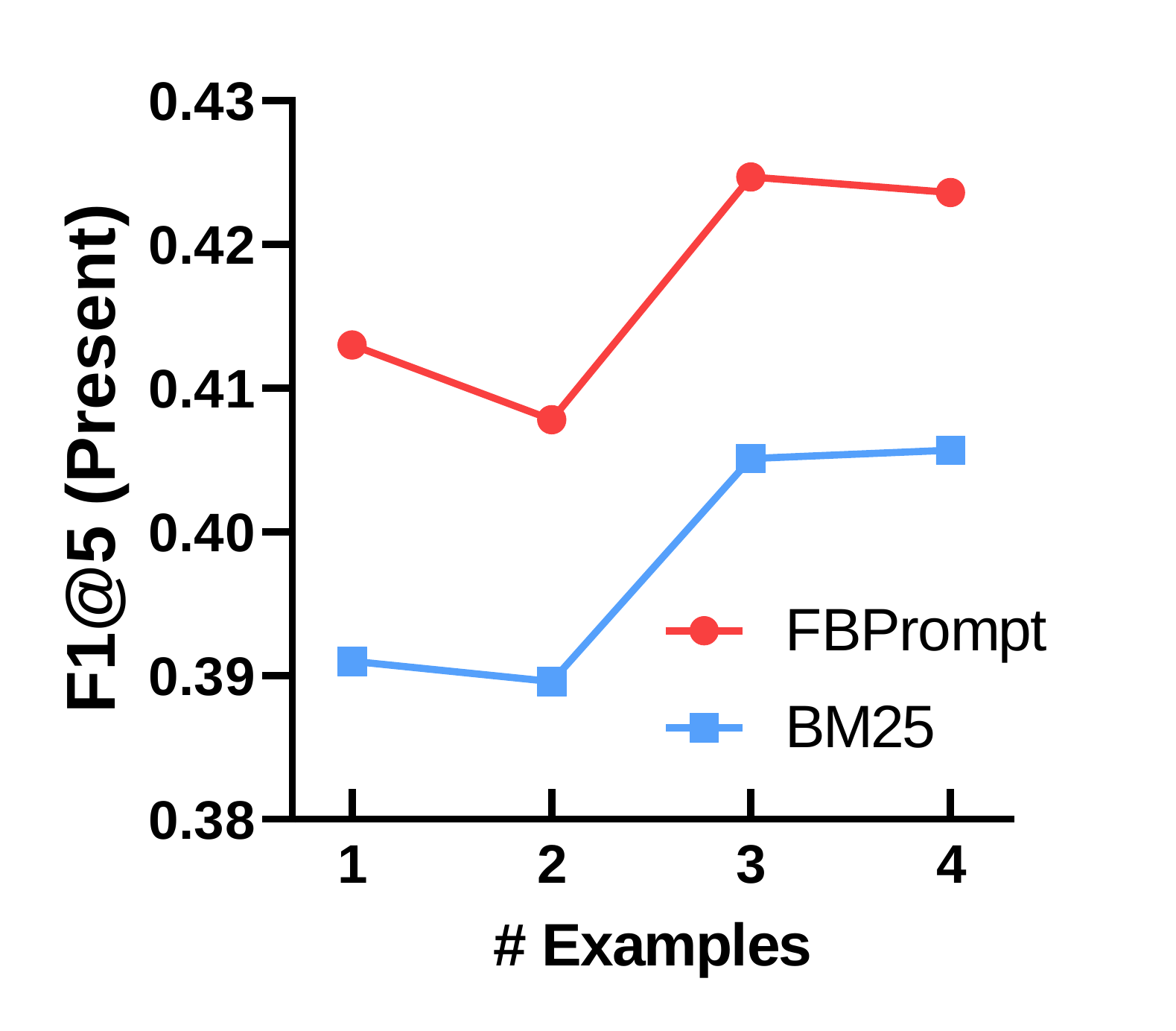}}
  \caption{Results of FBPrompt and BM25 with different numbers of examples on four datasets.}
  \label{fig:exam_num}
\end{figure}

\subsection{Comparison with Random Feedback}

Then, we simulate feedback by randomly generating predicted answers to observe whether the improvement of FBPrompt is really brought about by our carefully designed feedback. For the labeled answers set $\mathcal{A}^E_i$ from demonstration example $[D^E_i, Q^E_i, \mathcal{A}^E_i]$, we randomly selected a number $\hat{n}_1$ in the range $[0, |\mathcal{A}^E_i|\}]$, and randomly sampled $\hat{n_1}$ positive answers from the labeled answers set $\mathcal{A}^E_i$ as pseudo positive predicted answers $\mathcal{A}^\text{Pos}$. Similarly, we randomly selected a number $\hat{n}_2$ in the range $[0, |\mathcal{A}^E_i|\}]$, and randomly sampled $\hat{n_2}$ spans from the document $D^E_i$ as pseudo negative predicted answers $\mathcal{A}^\text{Neg}$. Then, we merged $\mathcal{A}^\text{Pos}$ and $\mathcal{A}^\text{Neg}$ as the pseudo predicted answers and executed FBPrompt to generate answers.

As shown in Table~\ref{table:noise_study}, the performance of FBPrompt drops significantly when random feedback is used, which shows that our constructed feedback is useful.

\begin{table}[t]
\centering
\caption{Comparsion with random feedback.}
\label{table:noise_study}
\begin{tabular}{lcc|cc|cc|cc|cc}

\Xhline{3\arrayrulewidth}
\multirow{3}{*}{} & \multicolumn{2}{c|}{MultiSpanQA} & \multicolumn{2}{c|}{QUOREF} & \multicolumn{2}{c|}{DROP} & \multicolumn{4}{c}{INSPEC} \\
\cline{2-11}
& \multirow{2}{*}{EM} & \multirow{2}{*}{PM} & \multirow{2}{*}{EM$_\text{G}$} & \multirow{2}{*}{F1} & \multirow{2}{*}{EM$_\text{G}$} & \multirow{2}{*}{F1} & \multicolumn{2}{c|}{Present} & \multicolumn{2}{c}{Absent}\\
\cline{8-11}
& &&&&&& F1@5 & F1@M & F1@5 & F1@M \\
\Xhline{2\arrayrulewidth}

FBPrompt & \textbf{64.60} & \textbf{83.11} & \textbf{73.60} & \textbf{80.55} & \textbf{62.00} & \textbf{69.11} & \textbf{0.425} & \textbf{0.499} &\textbf{0.034} & \textbf{0.055} \\

Random feedback & 60.95 & 81.40 & 64.40 & 72.11 & 60.72 & 68.03 &
0.415 & 0.482 & 0.029 & 0.047\\

\Xhline{3\arrayrulewidth}

\end{tabular}
\end{table}










\subsection{Number of Demonstration Examples}
We study whether FBPrompt exhibits consistent effectiveness when the number of demonstration examples varies. In Figure~\ref{fig:exam_num}, we report the changing trend of FBPrompt and BM25 when the number of examples changes from 1 to 4. We observe that with a varying number of examples in the four datasets, the performance of FBPrompt is consistently higher than that of BM25. Especially in the case of one-shot, FBPrompt largely outperforms BM25.

\subsection{Case Study}
\begin{table}

\scriptsize
\caption{Case Study}
\label{tb:case_study}
\begin{adjustbox}{center}
\renewcommand\arraystretch{1.1}
\begin{tabular}{p{6cm}|p{6cm}}
\Xhline{3\arrayrulewidth}
\makecell[c]{Demonstration Examples} & \makecell[c]{Test Question} \\
\hline
\textbf{Document:} \textbf{\textcolor{teal}{Glycogen}} functions as one of two forms of long - term energy reserves , with the other form being \textbf{\textcolor{teal}{triglyceride}} stores in adipose tissue ( i.e. , body fat ) . In humans , glycogen is made and stored primarily in the cells of the \textbf{\textcolor{red}{liver}} and \textbf{\textcolor{red}{skeletal muscle}} ...
& \textbf{Document:} Aflatoxins are poisonous carcinogens that are produced by certain molds ( \textbf{\textcolor{red}{Aspergillus flavus}} and \textbf{\textcolor{red}{Aspergillus parasiticus}} ) which grow in \textbf{\textcolor{teal}{soil , decaying vegetation , hay , and grains}} ...\\

\textbf{Question:} Which two forms of energy do muscles produce ? &
\textbf{Question:} Where are the organisms that produce aflatoxins found ?\\

\textbf{Gold Answers:} \textbf{\textcolor{teal}{Glycogen, triglyceride}} &
\textbf{Gold Answers:} \textbf{\textcolor{teal}{soil, decaying vegetation, hay, grains}}\\

\textbf{Feedback:\hspace{14em}} & \textbf{Baseline Predictions:} \textbf{\textcolor{red}{Aspergillus parasiticus, Aspergillus flavus}}\\
\textbf{Incorrect Answers:} \textbf{\textcolor{red}{liver, skeletal muscle}} \textbf{Answers Missed:} Glycogen, triglyceride & \textbf{FBPrompt Predictions:} \textbf{\textcolor{teal}{soil, decaying vegetation, hay, grains}}\\
\Xhline{3\arrayrulewidth}
\end{tabular}
\end{adjustbox}
\end{table}

A real case from MultiSpanQA is presented in Table~\ref{tb:case_study}. The left part shows an demonstration example for the test question in the right part. We can observe that the prediction of the baseline method (BM25) makes a mistake, since LLM observes `produce' in the question and directly finds answers around `produced', instead of analyzing the meaning of the question thoroughly. As for FBPrompt, our off-the-shelf model also observes `produce' in the question, and mistakenly finds the answers `liver', `skeletal musc' in the original document near `made', which is semantically close to `produce'. But after given a feedback, LLM learns not to be confused by such specific word, and tries to understand the entire question. Therefore, FBPrompt finally generates correct answers.


\section{Conclusion}
In this paper, we explore the performance of LLM in multi-span question answering, finding that existing in-context learning methods under-utilize labeled answers.  To alleviate this problem, we propose a novel prompting strategy called FBPrompt, which constructs and employs answer feedback from an off-the-shelf model to enhance in-context learning. Experiments on multiple datasets show that FBPrompt using answer feedback significantly improves the performance of LLM on MSQA tasks. In the future, we will deeply analyze the working principle of answer feedback, and try to integrate more 
useful feedback information into LLM for various tasks.

\subsubsection{Acknowledgements}
This work was supported in part by the NSFC (62072224) and in part by the CAAI-Huawei MindSpore Open Fund.

\bibliography{reference}
\bibliographystyle{splncs04}

\end{document}